\documentclass[conference]{IEEEtran}
\IEEEoverridecommandlockouts
\usepackage{cite}
\usepackage{amsmath,amssymb,amsfonts}
\usepackage{algorithmic}
\usepackage{graphicx}
\usepackage{textcomp}
\usepackage{xcolor}
\usepackage{subfigure}
\usepackage{hyperref}

\def\BibTeX{{\rm B\kern-.05em{\sc i\kern-.025em b}\kern-.08em
    T\kern-.1667em\lower.7ex\hbox{E}\kern-.125emX}}
\begin{document}

\title{AGSTN: Learning Attention-adjusted Graph Spatio-Temporal Networks for Short-term Urban Sensor Value Forecasting
}

\author{
\IEEEauthorblockN{Yi-Ju Lu}
\IEEEauthorblockA{\textit{Department of Statistics} \\
\textit{National Cheng Kung University}\\
Tainan, Taiwan \\
l852888@gmail.com}
\and
\IEEEauthorblockN{Cheng-Te Li}
\IEEEauthorblockA{\textit{Institute of Data Science} \\
\textit{National Cheng Kung University}\\
Tainan, Taiwan \\
chengte@mail.ncku.edu.tw}
}

\maketitle

\begin{abstract}
Forecasting spatio-temporal correlated time series of sensor values is crucial in urban applications, such as air pollution alert, biking resource management, and intelligent transportation systems. While recent advances exploit graph neural networks (GNN) to better learn spatial and temporal dependencies between sensors, they cannot model time-evolving spatio-temporal correlation (STC) between sensors, and require pre-defined graphs, which are neither always available nor totally reliable, and target at only a specific type of sensor data at one time. Moreover, since the form of time-series fluctuation is varied across sensors, a model needs to learn fluctuation modulation. To tackle these issues, in this work, we propose a novel GNN-based model, \textit{Attention-adjusted Graph Spatio-Temporal Network} (AGSTN). In AGSTN, multi-graph convolution with sequential learning is developed to learn time-evolving STC. Fluctuation modulation is realized by a proposed attention adjustment mechanism. Experiments on three sensor data, air quality, bike demand, and traffic flow, exhibit that AGSTN outperforms the state-of-the-art methods.
\end{abstract}

\begin{IEEEkeywords}
graph neural network, spatio-temporial correlation, time series, attention adjustment, urban computing, air quality, traffic flow, bike demand
\end{IEEEkeywords}


















\maketitle

\section{Introduction}
With the technologies of sensor networks and the Internet of Things, sensors are widely and geographically deployed in modern urban areas. Sensors established for different purposes collectively monitor the physical-world environment and continuously generate time-series data. Sensor time-series values are usually correlated in both spatial and temporal aspects. We consider that urban sensor time-series data possesses two properties, \textit{spatio-temporal correlation} and \textit{sequential effect}. For example, air-quality values of near-by monitoring stations can influence each other due to topography or wind. Traffic-flow values of sensors along the communication lines can rise and fall one after another. In addition, the time-series values of a sensor are sequentially self-correlated. Sensor values of air quality and traffic flow at the next time step are affected by those in the past few steps.  

Forecasting spatio-temporally correlated sensor values is a crucial task in various urban applications. Accurate forecasting of air quality allows people to manage themselves to avoid outdoor activities~\cite{accuair}. Precise prediction of bike demand enables better resource deployment and utilization~\cite{bikesp18}. Besides, Intelligent Transportation System (ITS)~\cite{STResNet} requires accurate and reliable traffic-flow forecasting. While conventional statistic models (e.g., ARIMA) and machine learning methods (e.g., support vector regression) can model the sequential effect of a single time series, deep recurrent and convolutional neural network-based models~\cite{cltfp,deepst} better model either spatial or temporal correlation between sensors so that the performance can get improved. Recent advances target at devising graph neural network (GNN) based models~\cite{gnnsvy,GCNpaper} to simultaneously model the spatial and temporal correlation among sensors~\cite{dcrnn,geoman,astgcn,stmgcn}, and the performance gets further boosted.

Nevertheless, existing urban sensor value forecasting models have several limitations and face some challenges. First, although the spatio-temporal correlation between sensors are modeled~\cite{stg2vec,astgcn,stmgcn}, the influence of such correlation can evolve. For example, wind direction that can affect how air pollutants propagate is different across seasons. Vehicle flows that vary on weekdays and weekends can affect traffic-flow values. Hence, it is necessary to further model the \textit{time-evolving spatio-temporal correlation}. Second, current graph neural network-based models~\cite{stg2vec,dgcnn,astgcn,stmgcn} require the pre-defined graphs (e.g., road networks, traffic structures of biking) as the model input. However, it is not realistic to assume that all sorts of sensors have pre-defined graphs or to presume that the existed graph structures are always available and in hand. For example, there are no connections between air-quality monitoring stations. Besides, modeling the spatio-temporal correlation based on the pre-defined graphs may not be able to capture potentially hidden mutual influence between sensors. Thus a practical and useful model is expected to accept \textit{no pre-defined graphs}. Third, existing models tend to target at specific sensor data, e.g., ST-GCN~\cite{stgcn} and AST-GCN~\cite{astgcn} for traffic flow, ST-MGCN~\cite{stmgcn} and STG2Vec~\cite{stg2vec} for bike demand, and DeepAir~\cite{deepair} and AccuAir~\cite{accuair} for air quality. While these data-specialized models are built to deal with the predictions of various urban sensors, we attempt to develop a \textit{general-purpose} model that can perform spatio-temporal forecasting for any sort of sensor data. Fourth, sensor time-series values can fluctuate over time. Due to the topography, the population, and the near-by types of point of interest (POI), the fluctuation's scale, frequency, and duration can vary across different sensors. How to perform fluctuation-aware modulation on the forecasted values is essential but challenging. However, none of the existing studies consider the learning of \textit{fluctuation modulation}. We aim at automatically learning the fluctuation modulation without using additional data but time series themselves.

Given the past time-series values of deployed sensors, this paper aims at forecasting sensor values at the next time steps. To overcome the abovementioned limitations and challenges, we propose a novel graph neural network-based model, \textbf{\underline{A}ttention-adjusted \underline{G}raph \underline{S}patio-\underline{T}emporal \underline{N}etwork} (AGSTN). The main idea of AGSTN is four-fold. First, we create multiple graphs to depict spatio-temporal correlation (STC) at the past time steps, and exploit graph neural networks to learn sensor STC embeddings. Our AGSTN model requires no pre-defined graph structures for representation learning. Second, we learn the time-evolving STC features by applying 1-D convolutional neural networks to STC embeddings. Third, we learn attention weights from the original time series for different sensors, and perform \textit{attention adjustment} on the raw predictions so that the fluctuation-aware modulation can be realized to make the forecasting values be within every sensor's reasonable range. Fourth, we leverage intrinsic mode functions (IMF) to generate additional features based on the original time series.

Below we summarize the contributions of this work.
\begin{itemize}
\item We identify several key issues in forecasting spatio-temporal correlated urban sensor values, including modeling time-evolving spatio-temporal correlation, no pre-defined graph structures, the model generality for all sorts of sensor data, and the learning of fluctuation modulation.
\item We develop a general-purpose graph neural network-based model, Attention-adjusted Graph Spatio-Temporal Network (AGSTN)
\footnote{Code can be accessed at \url{https://github.com/l852888/AGSTN}}, 
to deal with the identified key issues. The novelty lies in: (a) incorporating graph neural networks with sequential methods so that time-evolving STC can be learned, (b) requiring no pre-defined graphs, and (c) inventing the attention adjustment to fulfill the fluctuation modulation. 
\item Extensive experiments conducted on three different sensor data, including air quality, bike demand, and traffic flow, consistently exhibit that AGSTN outperforms state-of-the-art models in terms of not only lower error values but also higher ranking accuracy. 
\end{itemize}

This paper is organized as follows. 
We present the problem statement in Sec.~\ref{sec-prob}. Sec.~\ref{sec_method} describes the technical details of the proposed AGSTN. We report the experimental results in Sec.~\ref{sec-exp}, and conclude this work in Sec.~\ref{sec_conclude}.
\section{Problem Statement}
\label{sec-prob}
Let $S$ be the set of all sensors deployed in a specific geographical area, and $N=|S|$ be the number of sensors. Each sensor $s_i\in S$ is associated with a sequence of observed time-series real values, denoted by $\mathbf{x}_i$. We consider that the time-series data is recorded in a discrete manner, and thus denote $x_i^t$ as the observed value at time $t$. Similarly, we denote $\mathbf{X}$ and $\mathbf{X}^t\in \mathbb{R}^{1\times|S|}$ as the matrix of time-series values of all sensors, and their values at time $t$, respectively. 

\textbf{Short-term Sensor Value Forecasting (SSVF).} Given the observed time-series values of all sensors $S$ at previous $\tau$ time steps, denoted by a matrix $\left[\mathbf{X}^{t-\tau+1},\mathbf{X}^{t-\tau+2},\mathbf{X}^{t}\right]$, where $t$ is the current time step, the goal of SSVF problem is formulated as a next single-step spatio-temporal prediction given a fixed previous $\tau$-step observed values of all sensors as the input. Specifically, we aim at learning a function $f:\mathbb{R}^{N\times\tau} \rightarrow \mathbb{R}^{1\times N}$, that maps previous $\tau$ observations of all sensors to their predicted values at the next time steps, given by:
\begin{equation}
f(\left[\mathbf{X}^{t-\tau+1},\mathbf{X}^{t-\tau+2},\cdots,\mathbf{X}^{t}\right]) \rightarrow \mathbf{X}^{t+\Delta},
\end{equation}
where $\Delta$ is the number of short-term future time steps. In the experimemts, we will report the performance by varying different $\Delta$ values, i.e., $\Delta=1$ and $\Delta>1$, and varying the number of training time steps $\tau$.

In the SSVF problem, since we target at using short-term observations and aim to perform fine-grained time series prediction, we consider ``hour'' to be the time unit for air quality and bike demand data, and ``5 minutes'' to be the time unit for traffic data. We set the parameter $\tau$ to be a small integer (set $\tau=6$ by default). In other words, we make predictions by using the observed air-quality values, and bike-demand values in the past six hours; and the observed traffic-flow values in the past $30$ minutes.

\begin{figure}[!t]
\centering
\centerline{\includegraphics[width=1.0\linewidth]{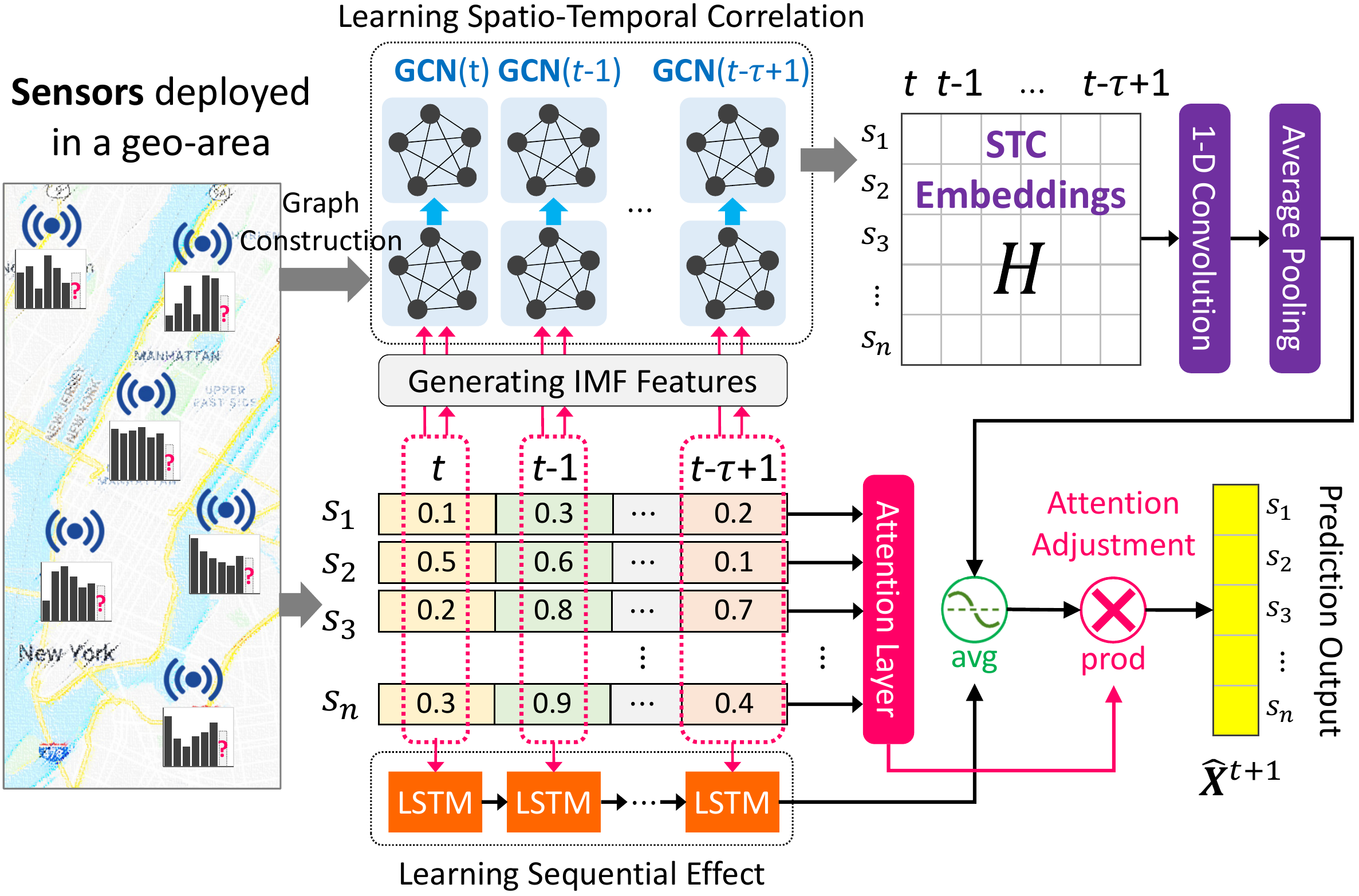} }
\caption{Overview of the proposed AGSTN.}
\label{fig:overview}
\end{figure}
\section{The Proposed AGSTN Model}
\label{sec_method}
We propose a novel model, \textbf{\underline{A}ttention-adjusted \underline{G}raph \underline{S}patio-\underline{T}emporal \underline{N}etwork} (AGSTN), to deal with the SSVF problem. The overview of AGSTN is shown in Figure~\ref{fig:overview}. AGSTN consists of five parts. First, we exploit \textit{Intrinsic Mode Functions} (IMF) to generate additional features from the original time series. Second, we aim at learning spatio-temporal correlation between sensors based on constructed graphs. \textit{Multi-graph convolutional network} (M-GCN) layers are incorporated to achieve the aim. Third, we use the sequential methods to make raw predictions from the original time series and the M-GCN-generated features, using recurrent and convolutional neural networks, respectively, from which time-evolving spatio-temporal correlation between sensors is modeled. Fourth, we impose an attention layer to learn the reasonable fluctuation tendency of every sensor. Last, by averaging raw predictions and applying the attention adjustment, the final prediction is generated.

\subsection{Generating IMF Features}
\label{subs_imf}
While the original time series are treated as features in our model, we generate additional features by decomposing the original ones. The Empirical Ensemble Decomposition (EEMD) algorithm~\cite{IMFS} is adopted. The idea is to use multi-granularity basic functions with variable amplitude and frequency over time to represent and approximate the original time series. By doing so, the model will be able to learn from time series with similar shapes and tendencies. The original time series are decomposed into intrinsic mode functions and the residuals. Given the observed time series $\mathbf{x}_i\in \mathbf{R}^{\tau\times 1}$ of sensor $s_i$, EEMD decomposes $\mathbf{x}_i$ to multiple intrinsic functions $\mathbf{IMF}_i=\{\mathbf{IMF}_1, \mathbf{IMF}_2,...,\mathbf{IMF}_K\}$, where $K$ is the number of intrinsic mode functions, and every $\mathbf{IMF}_k$ is $\tau$-dimensional. We treat every $\mathbf{IMF}_k$ as an additional time-series feature, and the $\mathbf{IMF}^{t_j}\in\mathbb{R}^{K \times N}$ be the IMF feature matrix of all sensors at time step $t_j$.

\subsection{Modeling Spatial-Temporal Correlation}
\label{subs_mgcn}
Sensor values can be correlated with each other due to a variety of spatio-temporal properties. Traffic-flow and bike-demand values in an urban area can propagate between places that people commute from and to over time. Air-quality values can exhibit some kind of spatial distributions and diffusion patterns according to the topography. We aim at modeling such kind of spatio-temporal correlation between sensors. The idea is constructing multiple graphs to represent all of the potential rise-or-fall co-influence between sensors. Then we leverage graph convolutional networks to learn the correlation.

\textbf{Multi-Graph Construction.} Since the spatial correlation between sensors can vary from one time to another, we construct a graph $\mathcal{G}^{t_j}=(S,E,\mathbf{A}^{t_j})$ for every time $t_j\in[t-\tau+1,\cdots,t]$, where $S$ is the set of sensor nodes, $E$ is the set of edges, and $\mathbf{A}^{t_j}$ is the set of edge weights. We consider every graph $\mathcal{G}^{t_j}$ is fully-connected, i.e., $e_{uv}\in E, \forall u,v\in S, u\neq v$. Sets $S$ and $E$ are the same for all graphs $\mathcal{G}^{t_j}$, but the sets of edge weights $\mathbf{A}^{t_j}$ are different. Edge weights are expected to depict the potential that the values of two sensors can be influenced by one another. Hence, we use cosine similarity for edge weights, i.e., $\omega_{uv}=\frac{\mathbf{x}_u\cdot \mathbf{x}_v}{\|\mathbf{x}_u\|\|\mathbf{x}_v\|} \in \mathbf{A}$, where $\mathbf{A}\in\mathbb{R}^{N\times N}$

\textbf{Graph Convolution Layer.} We aim to learn the spatio-temporal correlation (STC) embeddings for every sensor based on the multiple graphs constructed from the past $\tau$ time steps. We adopt graph convolutional networks (GCN)~\cite{GCNpaper} to derive STC embeddings of sensors. A GCN is a multi-layer neural network that performs on graph data and generates embedding vectors of nodes according to their neighborhoods. GCN is able to capture information from a node's direct and indirect neighbors through stacking layer-wise graph convolution. We learn a GCN for each constructed graph $\mathcal{G}^{t_j}$ so as to model spatio-temporal correlation between sensors at every time step $t_j$. That is, we devise a multi-graph convolutional (MGC) network to generate STC embeddings. Given the matrix $\mathbf{A}^{t_j}$ for a graph $\mathcal{G}^{t_j}$, and $\mathbf{P}^{t_j}=[\mathbf{X}^{t_j},\mathbf{IMF}^{t_j}]$ depicting the matrix of feature vectors for sensor nodes in $\mathcal{G}_{t_j}$, the new $b$-dimensional node feature matrix $\mathbf{H}^{(l+1)}_{t_j}\in \mathbb{R}^{N\times b}$ can be derived by
\begin{equation}
\mathbf{H}^{(l+1)}_{t_j} = \rho(\tilde{\mathbf{A}}^{t_j}\mathbf{H}^{(l)}_{t_j}\mathbf{W}_l), t_j\in[t-\tau+1,\cdots,t]
\end{equation}
where $l$ is the layer number, $\tilde{\mathbf{A}}^{t_j}=\mathbf{D}^{-\frac{1}{2}}\mathbf{A}^{t_j}\mathbf{D}^{-\frac{1}{2}}$ is the normalized symmetric weight matrix ($\mathbf{D}_{ii}=\sum_q \mathbf{A}^{t_j}_{iq}$), and $\mathbf{W}_l\in\mathbb{R}^{b\times K}$ ($K$ is the number of IMF features) is the matrix of learnable parameters at the $l$-th GCN layer. $\rho$ is an activation function, i.e., a ReLU $\rho(x)=\text{max}(0,x)$. Here $\mathbf{H}^{(0)}_{t_j}$ is set to be $\mathbf{P}^{t_j}$. We choose to stack two GCN layers in derive the learned sensor representation, denoted as $\mathbf{H}_{t_j}\in \mathbb{R}^{N\times b}$. We set $b=1$ since our goal is to predict the next single-step sensor value. By concatenating $\mathbf{H}_{t_j}$ over all past $\tau$ time steps, we can have the STC embeddings $\mathbf{H}\in \mathbb{R}^{N\time\tau}$ for all sensors.

\subsection{Raw Predictions by Sequential Models}
\label{subs_rcnn}
Sensor time-series values are sequentially evolved. Hence it is natural to exploit sequential models to model time-evolving spatio-temporal correlation between sensors and generate the raw predictions. We consider two sequential models, recurrent neural networks (RNN) and convolutional neural networks (CNN), to model temporal dynamics from two aspects. RNN is used to capture the global evolution tendency based on the original time series of all sensors. CNN is applied to STC sensor embeddings so that the propagation from the local geo-spatial neighborhood along past time steps can be modeled. Both RNN and CNN can produce raw next-step predictions.

\textbf{RNN-based Raw Prediction.} Given the sequence of original observations in $N$ monitor stations at past $\tau$ time steps $[\mathbf{X}^{t-\tau+1},\cdots,\mathbf{X}^{t}]$, we utilize Long Short-Term Memory (LSTM)~\cite{lstm97} to learn the global temporal dynamics. Each LSTM state has two inputs, the observations $\mathbf{X}^{t}$ of all sensors at the current time step $t$ and the previous state's output vector $\mathbf{h}^{t-1}$. The LSTM-based representation learning can be depicted by:
\begin{equation}
\mathbf{h}^{t_j}=LSTM(\mathbf{X}^{t_j}),
\end{equation}
where $t_j\in [t-\tau+1,\cdots,t]$, in which $\tau$ is the number of LSTM states. By feeding the last LSTM output vector $\mathbf{h}^t$ into a fully-connected layer, we generate the RNN-based raw prediction $\hat{\mathbf{r}}^{t+1}\in \mathbb{R}^{N}$.

\textbf{CNN-based Raw Prediction.} We utilize 1-D convolution neural network to learn the propagation influence based on the derived STC sensor embeddings $\mathbf{H}\in \mathbb{R}^{N\times\tau}$. We consider $\lambda$ consecutive time steps at one time to model their sequential information produced by GCN. An independent 1-D CNN is applied on each sensor. Hence the filter is set as $\mathbf{W}_f\in \mathbb{R}^{1\times \lambda}$. Then the output vector $\mathbf{C}\in \mathbb{R}^{N\times (\tau+\lambda-1)}$ is:
\begin{equation}
\mathbf{C}=\text{ReLU}(\mathbf{W}_{f}\cdot \mathbf{H}_{\tau:\tau+\lambda-1}+b_{f}))
\end{equation}
where $\mathbf{W}_{f}$ is the matrix of learnable parameters, $ReLU$ is the activation function, $\mathbf{H}_{\tau:\tau+\lambda-1}^{l+1}$ depicts sub-matrices whose first row's index is from $\tau$ to $\tau+\lambda-1$, and $b_f$ is the bias term. We generate the CNN-based raw prediction through applying average pooling to $\mathbf{C}$ and obtain $\hat{\mathbf{c}}^{t+1}\in \mathbb{R}^{N}$.

\subsection{Learning Attention Weights for Adjustment}
\label{subs_att}
Since sensors are deployed at different geographical positions, the scale, frequency, and duration of their fluctuation on sensor values can be varied. In addition, some sort of sensor values can change a lot (e.g., traffic flow) within a short-term time period (e.g., 6 hours), and another can have quite minor variation (e.g., air quality). To make the model capture the basic shape and fluctuation tendency of sensors, we aim to learn attention weights from their past time series, which is used as a scaling factor of the predicted values. The derived attention weight for each sensor will be used to adjust the raw predictions and generate the final results.

The learning and use of attention weights consist of three phases. First, we average the raw CNN-based and RNN-based predictions of all sensors, $\hat{\mathbf{r}}^{t+\Delta}$ and $\hat{\mathbf{c}}^{t+\Delta}$, where $\Delta$ is the number of future time steps, to be the ensemble prediction, given by:
\begin{equation}
\hat{\mathbf{X}}^{t+\Delta}_{ens}=average(\hat{\mathbf{r}}^{t+\Delta},\hat{\mathbf{c}}^{\Delta+1}).
\end{equation}
Second, we learn the attention weights $\mathbf{a}\in\mathbb{R}^{1\times N}$ based on the original observations of past $\tau$ time steps. Then we apply an activation function to generate $\mathbf{a}$. This can be depicted by:
\begin{equation}
\mathbf{a}=\sigma(\mathbf{W}_a\cdot[\mathbf{X}^{t-\tau+1},\mathbf{X}^{t-\tau+2},\cdots,\mathbf{X}^{t}]),
\end{equation}
where $\mathbf{W}_a\in \mathbb{R}^{1\times N}$ is the learnable weight matrix, $\sigma$ is the \textit{sigmoid} function. Third, the learned attention weights $\mathbf{a}$ is used to adjust the ensembled prediction $\hat{\mathbf{X}}^{t+\Delta}_{ens}$ and generate the final prediction $\hat{\mathbf{X}}^{t+\Delta}$ through element-wise product, given by:
\begin{equation}
\hat{\mathbf{X}}^{t+\Delta}=\mathbf{a}\odot \hat{\mathbf{X}}^{t+\Delta}_{ens},
\end{equation}
where $\odot$ is the element-wise Hadamard product. 

\subsection{Objective Function}
\label{subs_pred}
We choose to employ \textit{mean squared error} (MSE) to be the training objective of our AGSTN model. The objective function $\mathcal{L}$ can be written as:
\begin{equation}
\mathcal{L}(\Theta)=\frac{1}{N}\sum_{s\in S}(\mathbf{X}^{t+1}_s,\hat{\mathbf{X}}^{t+1}_s),
\end{equation}
where $\mathbf{X}^{t+1}_s$ is the ground-truth sensor value of sensor $s$ at the next time step $t+1$, and $\Theta$ is the set of all trainable parameters. The MSE loss function is minimized by back-propagation.

\section{Experiments}
\label{sec-exp}

\begin{table}[!t]
\centering
\caption{Statistics of three datasets.}
\label{tab:data}
\resizebox{\linewidth}{!}{%
\begin{tabular}{l|c|c|c}
\hline
 & Air Quality & Bike Demand & Traffic Flow \\ \hline\hline
Time span & 2018.01-12 & 2017.01-07 & 2012.03-06 \\ \hline
\# Records & 8,760 & 10,176 & 34,271 \\ \hline
\# Sensors & 26 & 827 & 207 \\ \hline
Location & Taiwan & New York & Los Angeles \\\hline
Time unit & hourly & 30 mins & 5 mins \\ \hline
\# Past steps ($\tau$) & 6 hrs & 6 hrs & 30 mins \\ \hline
\end{tabular}%
}
\end{table}

\begin{table*}[!t]
\centering
\caption{Main results on the next single time-step forecasting (i.e., $\Delta=1$). MSE and RMSE are error values, and P@5 and NDCG are ranking accuracy. The best model and the best competitor are highlighted by \textbf{bold} and \underline{underline}, respectively. The last row ``Improvement'' is computed by $\frac{|\text{bold}-\text{underline}|}{\text{underline}}\times100\%$.}
\label{tab:overallexp}
\resizebox{\textwidth}{!}{%
\begin{tabular}{l|cccc|cccc|cccc}
\hline
\multicolumn{1}{c}{}&\multicolumn{4}{c}{\textbf{Air Quality}}&\multicolumn{4}{c}{\textbf{Bike Demand}}&\multicolumn{4}{c}{\textbf{Traffic Flow}}\\\hline
\multicolumn{1}{c|}{}&\multicolumn{2}{c}{Error}&\multicolumn{2}{c|}{Ranking}&\multicolumn{2}{c}{Error}&\multicolumn{2}{c|}{Ranking}&\multicolumn{2}{c}{Error}&\multicolumn{2}{c}{Ranking}\\\hline
Model&MAE&RMSE&P@5&NDCG&MAE&RMSE&P@5&NDCG&MAE&RMSE&P@5&NDCG\\\hline\hline
ARIMA&11.2737&46.0466&0.3387&0.3332&7.3051&10.4379&0.3054&0.2432&6.6607&18.4607&0.3542&0.3254\\
SVR&10.9249&37.8123&0.3485&0.3483&4.7513&8.0558&0.3472&0.2638&3.3812&8.9780&\underline{0.4596}&\underline{0.5982}\\
FC-LSTM&8.8307&18.4758&0.3933&0.4321&6.5288&7.7125&0.4080&0.3373&3.1870&4.4329&0.4305&0.5479\\
DCRNN&9.0171&18.4077&0.3822&0.425&2.8483&3.6403&0.5227&\underline{0.5566}&2.7431&3.8882&0.4389&0.5593\\
AST-GCN&8.6277&18.4002&\underline{0.4134}&\underline{0.4716}&3.2184&4.0919&\underline{0.5315}&0.5182&\underline{2.4462}&\underline{3.3060}&0.4493&0.5831\\
ST-MGCN&\underline{8.2827}&\underline{17.8944}&0.3994&0.4427&\underline{2.8381}&\underline{3.6400}&0.5290&0.5560&2.8043&4.0127&0.4399&0.5604\\\hline
\textbf{AGSTN-imf}&7.0258&16.7142&0.5947&0.6366&2.8176&3.6594&0.5210&0.5516&2.3177&3.1401&0.4848&0.5993\\
\textbf{AGSTN}& \textbf{6.9273}& \textbf{16.6405}& \textbf{0.6007}& \textbf{0.6423}& \textbf{2.7464}& \textbf{3.5380}& \textbf{0.5387}& \textbf{0.5634}& \textbf{2.2429}& \textbf{3.1032}& \textbf{0.4962}& \textbf{0.6125}\\\hline
\textbf{Improvement}& \textbf{16.3\%}& \textbf{7.0\%}&\textbf{45.3\%}&\textbf{36.2\%}& \textbf{3.2\%}& \textbf{2.8\%}&\textbf{1.6\%}&\textbf{1.2\%}& \textbf{8.3\%}& \textbf{6.1\%}&\textbf{8.2\%}&\textbf{2.4\%}\\\hline
\end{tabular}}
\end{table*}

\begin{table*}[!t]
\centering
\caption{Results on the forecasting of the next five time-steps (i.e., $\Delta=5$). 
}
\label{tab:overallexp_5}
\resizebox{\textwidth}{!}{%
\begin{tabular}{l|cccc|cccc|cccc}
\hline
\multicolumn{1}{c}{}&\multicolumn{4}{c}{\textbf{Air Quality}}&\multicolumn{4}{c}{\textbf{Bike Demand}}&\multicolumn{4}{c}{\textbf{Traffic Flow}}\\\hline
\multicolumn{1}{c|}{}&\multicolumn{2}{c}{Error}&\multicolumn{2}{c|}{Ranking}&\multicolumn{2}{c}{Error}&\multicolumn{2}{c|}{Ranking}&\multicolumn{2}{c}{Error}&\multicolumn{2}{c}{Ranking}\\\hline
Model&MAE&RMSE&P@5&NDCG&MAE&RMSE&P@5&NDCG&MAE&RMSE&P@5&NDCG\\\hline\hline
FC-LSTM&14.0541&23.8502&0.3356&0.2006&6.5833&7.7567&0.4124&0.3353&5.2397&6.4976&0.3452&0.4698\\
DCRNN&9.7865&19.6419&0.3893&0.4197&\textbf{3.3658}&\textbf{4.3084}&\underline{0.5048}&0.5143&5.0858&6.3328&0.4070&0.5414\\
AST-GCN&10.3618&20.3901&0.3737&0.3908&4.5191&5.6869&0.4519&0.4018&5.0742&6.2875&0.3947&0.5274\\
ST-MGCN&10.009&19.7600&0.3848&0.4182&4.5876&5.7613&0.4248&0.3920&4.8635&6.1037&0.4400&0.5641\\\hline
\textbf{AGSTN-imf}&\textbf{9.3670}&\textbf{19.3148}&\textbf{0.4378}&\textbf{0.4692}&\underline{3.4077}&\underline{4.3598}&0.4819&\underline{0.5157}&\underline{4.7689}&\underline{5.9536}&\underline{0.4638}&\underline{0.5906}\\
\textbf{AGSTN}& \underline{9.4171}& \underline{19.4865}& \underline{0.4320}& \underline{0.4541}& 3.4933& 4.4395& \textbf{0.5148}&\textbf{0.5298}&\textbf{4.7244}&\textbf{ 5.9297}&\textbf{0.4639}&\textbf{0.5950} \\\hline
\end{tabular}}
\end{table*}

\label{sec_exp}

\subsection{Dataset and Settings}
\label{subs_expset}
\textbf{Datasets.} Three datasets of spatio-temporal correlated time series are employed in our experiments. The types of time-series values include air quality, bike demand, and traffic flow. We provide data statistics in Table~\ref{tab:data}. For air-quality data, we collect hourly PM2.5 data of the northern region from the Environmental Protection Agency in Taiwan~\footnote{https://taqm.epa.gov.tw/taqm/en/default.aspx}. For bike-demand data, we make use of the Citi Bike public dataset in New York City~\footnote{https://www.citibikenyc.com/system-data}, whose sensors collect the demand number of bikes every 30 minutes. For traffic-flow data, we utilize the METR-LA dataset~\footnote{https://github.com/liyaguang/DCRNN}, which contains traffic information collected from loop detectors in the highway of Los Angeles. The sensor readings are aggregated into 5-minutes windows.

\textbf{Competing Methods.} We compare the proposed AGSTN with a number of baselines and state-of-the-art (SOTA) methods, as listed below. (1) \textbf{ARIMA}: Auto-Regressive Integrated Moving Average model~\footnote{https://www.statsmodels.org/stable/index.html} is widely used in time series prediction. (2) \textbf{SVR}: Support Vector Regression~\footnote{https://scikit-learn.org/stable/} uses linear support vector machine for the regression prediction. (3) \textbf{FC-LSTM}~\cite{seq2seq}: Recurrent Neural Network with fully connected LSTM hidden units. (4) \textbf{DCRNN}~\cite{dcrnn}: A graph convolution-based model to learn the spatio-temporal dependency by integrating graph convolution into the gated recurrent unit. (5) \textbf{AST-GCN}~\cite{astgcn}: one of the SOTA models, it is an attention-based graph convolution model to learn the spatial and temporal dependencies with convolution structures. (6) \textbf{ST-MGCN}~\cite{stmgcn}: one of the SOTA models, it is a multi-graph convolution-based model to learn the spatial dependency and temporal correlation using the contextual gated recurrent neural network. The hyperparapameters of competing methods are set by following the respective studies. For some competing methods (i.e., AST-GCN and ST-MGCN) need pre-defined input graphs, such as road network and transportation graph, to have a fair comparison, we use our graph construction method mentioned in Section~\ref{subs_mgcn}.

\textbf{Evaluation Metrics.} To measure how the predicted values are close to the ground truth, we adopt \textit{Mean Absolute Error} (MAE) and \textit{Rooted Mean Squared Error} (RMSE) as the evaluation metrics. Besides, it is also essential to identify sensors with the highest values in the next time step. Hence, by sorting sensors based on their forecasted values in a descending manner, we also consider it as a ranking problem. The metrics, \textit{Precision}$@k$ (P@$k$) ($k=5$ by default) and \textit{Normalized Discounted Cumulative Gain} (NDCG)~\cite{mir99}, are employed. For the first two metrics, lower scores indicate better performance. For the latter two, higher is better.

\textbf{Evaluation Settings.} We follow the settings of SOTA models AST-GCN~\cite{astgcn} and ST-MGCN~\cite{stmgcn} to determine the training, validation, and testing sets. We split each dataset in a chronological order with $70$\% for training, $10$\% for validation, and $20$\% for testing. The proposed model is termed \textbf{AGSTN}. Its batch size is $32$, and the initial learning rate is $10^{-3}$ with a decay rate of $0.7$ after every $5$ epochs. The 1-D CNN filter size is $3$. To understand how the proposed AGSTN itself can perform, we create a variation without using additional IMF features, termed \textbf{AGSTN-imf}. To have a fair comparison, for all models, we set the number of training epochs to be $200$, and stop training if the validation loss does not decrease for $10$ epochs.

\subsection{Experimental Results}
\label{subs_expres}
\textbf{Next-Step Forecasting.} The results on predicting the sensor values of the next step ($\Delta=1$) are shown in Table~\ref{tab:overallexp}. We can observe that the proposed AGSTN consistently leads to the best performance across three datasets under four evaluation metrics. The improvement is on average 26\% for air quality, 2.2\% for bike demand, and 6.3\% for traffic flow. By looking into the results, we further have the following four findings. First, even without adding the IMF features (i.e., AGSTN-IMF), our model still apparently outperforms the SOTA models and baselines. Such results exhibit the promising prediction capability of the proposed AGSTN model itself. Second, the powerful representation capability of graph convolutional networks leads to better performance of the SOTA methods AST-GCN and ST-MGCN, and our AGSTN. Nevertheless, our AGSTN outperforming the SOTA models further proves the usefulness of learning from fully-connected (non-predefined) graphs, attention adjustment, and incorporating IMF features. We think unrestricted graph construction can better model the correlation and dependency between sensors, comparing to the SOTA models that require pre-defined graphs of connected grids, traffic, and transportation. Third, the proposed AGSTN is able to not only generate predictions with lower errors, but be capable of accurately identifying sensors with higher sensor values in the ranking settings. Fourth, among the three datasets, the superiority of our AGSTN is significant on air-quality data and limited on bike-demand data. We think the reason could be the demand for bikes is more complicated and influenced by more factors, including meteorology and nearby point-of-interest properties, in addition to only historical values, which is supported by some recent studies~\cite{bikesp18,stg2vec}.

\textbf{Multi-Step Forecasting.} Generally, in making decisions, it will be more helpful to accurately predict the sensor values at multiple next steps. We conduct another experiment to examine whether the proposed AGSTN can also generate satisfying forecasting performance of multiple future steps by setting $\Delta=5$. 
In other words, the prediction setting can be depicted as: $f([\mathbf{X}^{t-\tau+1},\mathbf{X}^{t-\tau+2},\cdots,\mathbf{X}^{t}]) \rightarrow \mathbf{X}^{t+5}$. The results are shown in Table\ref{tab:overallexp_5}. 
We can have the following findings. First, our AGSTN and its variant AGSTN-imf can still outperform the competing methods across three datasets and four metrics, especially in the datasets of Air Quality and Traffic Flow. Although, in some cases, i.e., the error measures in Bike Demand data, AGSTN or AGSTN-imf have a bit higher MAE and RMSE scores than DCRNN, their differences are quite small. Second, 
by looking into the results, we find that in forecasting the more time steps, the IMF features seem not to be so helpful for the prediction. These features may be more suitable for the prediction of less future time steps. Third, despite the proposed AGSTN cannot always have the lowest prediction errors of sensor values, it leads to the best performance on being capable of accurately identifying sensors with higher sensor values at multiple future steps in the metrics of ranking and recommendation (P@5 and NDCG).

\section{Conclusions and Future Work}
\label{sec_conclude}
In this paper, we propose a novel graph neural network-based model, Attention-adjusted Graph Spatio-Temporal Network (AGSTN), to forecast short-term sensor values. AGSTN requires only the original sensor time series and needs no pre-defined graphs, and thus can be applied for any sort of urban sensor data, such as air quality, taxi demand, and crowd flow. The technical novelty of AGSTN lies in incorporating multi-graph convolution with sequential learning to capture time-evolving spatio-temporal correlation between sensors. And AGSTN learns attention weights from the original time series to adjust and modulate the raw prediction. Experiments on three real-world datasets show that the forecasting performance of AGSTN is superior to state-of-the-art models.

In fact, various urban sensor time-series values are usually affected by multiple factors, such as near-by points of interest, meteorology, topography, and the population. Ongoing work is to consider multi-source data as additional features of AGSTN so that more external influencing factors can be modeled. In addition, while the current graph construction is based on past time-series similarity between sensors, we are seeking to perform the learning of graph structures by adding it into the objective function. By doing so, the learned graph will be able to explicitly interpret which sensors are correlated each other over time.

\vspace{-5pt}

\section*{Acknowledgments}
This work is supported by Ministry of Science and Technology (MOST) of Taiwan under grants 109-2636-E-006-017 (MOST Young Scholar Fellowship) and 109-2221-E-006-173, and also by Academia Sinica under grant AS-TP-107-M05.


\bibliographystyle{IEEEtranS}
\bibliography{myrefs}

\begin{thebibliography}{10}
\providecommand{\url}[1]{#1}
\csname url@samestyle\endcsname
\providecommand{\newblock}{\relax}
\providecommand{\bibinfo}[2]{#2}
\providecommand{\BIBentrySTDinterwordspacing}{\spaceskip=0pt\relax}
\providecommand{\BIBentryALTinterwordstretchfactor}{4}
\providecommand{\BIBentryALTinterwordspacing}{\spaceskip=\fontdimen2\font plus
\BIBentryALTinterwordstretchfactor\fontdimen3\font minus
  \fontdimen4\font\relax}
\providecommand{\BIBforeignlanguage}[2]{{%
\expandafter\ifx\csname l@#1\endcsname\relax
\typeout{** WARNING: IEEEtranS.bst: No hyphenation pattern has been}%
\typeout{** loaded for the language `#1'. Using the pattern for}%
\typeout{** the default language instead.}%
\else
\language=\csname l@#1\endcsname
\fi
#2}}
\providecommand{\BIBdecl}{\relax}
\BIBdecl

\bibitem{mir99}
R.~A. Baeza-Yates and B.~Ribeiro-Neto, \emph{Modern Information
  Retrieval}.\hskip 1em plus 0.5em minus 0.4em\relax USA: Addison-Wesley
  Longman Publishing Co., Inc., 1999.

\bibitem{dgcnn}
Z.~Diao, X.~Wang, D.~Zhang, Y.~Liu, K.~Xie, and S.~He, ``Dynamic
  spatial-temporal graph convolutional neural networks for traffic
  forecasting,'' in \emph{Proceedings of The Thirty-Third AAAI Conference on
  Artificial Intelligence}, ser. AAAI '19, 2019, pp. 890--897.

\bibitem{stmgcn}
X.~Geng, Y.~Li, L.~Wang, L.~Zhang, Q.~Yang, J.~Ye, and Y.~Liu, ``Spatiotemporal
  multi-graph convolution network for ride-hailing demand forecasting,'' in
  \emph{Proceedings of the AAAI Conference on Artificial Intelligence}, 2019,
  pp. 3656--3663.

\bibitem{astgcn}
S.~Guo, Y.~Lin, N.~Feng, C.~Song, and H.~Wan, ``Attention based
  spatial-temporal graph convolutional networks for traffic flow forecasting,''
  \emph{Proceedings of the AAAI Conference on Artificial Intelligence},
  vol.~33, pp. 922--929, 07 2019.

\bibitem{lstm97}
S.~Hochreiter and J.~Schmidhuber, ``Long short-term memory,'' \emph{Neural
  Comput.}, vol.~9, no.~8, pp. 1735--1780, 1997.

\bibitem{IMFS}
N.~E. Huang, Z.~Shen, S.~R. Long, M.~C. Wu, H.~H. Shih, Q.~Zheng, N.-C. Yen,
  C.~C. Tung, and H.~H. Liu, ``The empirical mode decomposition and the hilbert
  spectrum for nonlinear and non-stationary time series analysis,''
  \emph{Proceedings of the Royal Society of London. Series A: Mathematical,
  Physical and Engineering Sciences}, vol. 454, no. 1971, pp. 903--995, 1998.

\bibitem{bikesp18}
P.~Hulot, D.~Aloise, and S.~D. Jena, ``Towards station-level demand prediction
  for effective rebalancing in bike-sharing systems,'' in \emph{Proceedings of
  the 24th ACM SIGKDD International Conference on Knowledge Discovery and Data
  Mining}, ser. KDD ’18, 2018, pp. 378--386.

\bibitem{GCNpaper}
\BIBentryALTinterwordspacing
T.~N. Kipf and M.~Welling, ``Semi-supervised classification with graph
  convolutional networks,'' \emph{CoRR}, vol. abs/1609.02907, 2016. [Online].
  Available: \url{http://arxiv.org/abs/1609.02907}
\BIBentrySTDinterwordspacing

\bibitem{dcrnn}
Y.~Li, R.~Yu, C.~Shahabi, and Y.~Liu, ``Diffusion convolutional recurrent
  neural network: Data-driven traffic forecasting,,'' in \emph{Proceedings of
  International Conference on Learning Representations}, ser. ICLR '17, 2017.

\bibitem{stg2vec}
Y.~Li, Z.~Zhu, D.~Kong, M.~Xu, and Y.~Zhao, ``Learning heterogeneous
  spatial-temporal representation for bike-sharing demand prediction,'' in
  \emph{Proceedings of The Thirty-Third AAAI Conference on Artificial
  Intelligence}, ser. AAAI '19, 2019, pp. 1004--1011.

\bibitem{geoman}
Y.~Liang, S.~Ke, J.~Zhang, X.~Yi, and Y.~Zheng, ``Geoman: Multi-level attention
  networks for geo-sensory time series prediction,'' in \emph{Proceedings of
  the Twenty-Seventh International Joint Conference on Artificial
  Intelligence}, ser. IJCAI '18.\hskip 1em plus 0.5em minus 0.4em\relax
  International Joint Conferences on Artificial Intelligence Organization,
  2018, pp. 3428--3434.

\bibitem{accuair}
Z.~Luo, J.~Huang, K.~Hu, X.~Li, and P.~Zhang, ``Accuair: Winning solution to
  air quality prediction for kdd cup 2018,'' in \emph{Proceedings of the 25th
  ACM SIGKDD International Conference on Knowledge Discovery and Data Mining},
  ser. KDD '19, 2019, pp. 1842--1850.

\bibitem{seq2seq}
I.~Sutskever, O.~Vinyals, and Q.~V. Le, ``Sequence to sequence learning with
  neural networks,'' in \emph{Proceedings of the 27th International Conference
  on Neural Information Processing Systems - Volume 2}, ser. NIPS '14, 2014, p.
  3104–3112.

\bibitem{cltfp}
Y.~Wu and H.~Tan, ``Short-term traffic flow forecasting with spatial-temporal
  correlation in a hybrid deep learning framework,'' 2016.

\bibitem{gnnsvy}
Z.~{Wu}, S.~{Pan}, F.~{Chen}, G.~{Long}, C.~{Zhang}, and P.~S. {Yu}, ``A
  comprehensive survey on graph neural networks,'' \emph{IEEE Transactions on
  Neural Networks and Learning Systems}, pp. 1--21, 2020.

\bibitem{deepair}
X.~Yi, J.~Zhang, Z.~Wang, T.~Li, and Y.~Zheng, ``Deep distributed fusion
  network for air quality prediction,'' in \emph{Proceedings of the 24th ACM
  SIGKDD International Conference on Knowledge Discovery and Data Mining}, ser.
  KDD ’18, 2018, pp. 965--973.

\bibitem{stgcn}
B.~Yu, H.~Yin, and Z.~Zhu, ``Spatio-temporal graph convolutional neural
  network: {A} deep learning framework for traffic forecasting,'' in
  \emph{Proceedings of the Twenty-Seventh International Joint Conference on
  Artificial Intelligence}, ser. IJCAI '18, 2018.

\bibitem{deepst}
J.~Zhang, Y.~Zheng, D.~Qi, R.~Li, and X.~Yi, ``Dnn-based prediction model for
  spatio-temporal data,'' in \emph{Proceedings of the 24th ACM SIGSPATIAL
  International Conference on Advances in Geographic Information Systems}, ser.
  SIGSPACIAL ’16, 2016.

\bibitem{STResNet}
J.~Zhang, Y.~Zheng, D.~Qi, R.~Li, X.~Yi, and T.~Li, ``Predicting citywide crowd
  flows using deep spatio-temporal residual networks,'' \emph{Artificial
  Intelligence}, vol. 259, pp. 147 -- 166, 2018.

\end{thebibliography}

\end{document}